\documentclass[10pt,twocolumn,letterpaper]{article}

\usepackage{iccv}
\usepackage{times}
\usepackage{epsfig}
\usepackage{graphicx}
\usepackage{amsmath}
\usepackage{amssymb}
\usepackage{bbding}

\usepackage[numbers,sort,compress]{natbib}
\usepackage{soul}
\usepackage[dvipsnames]{xcolor}
\usepackage{xspace}
\usepackage[normalem]{ulem}

\usepackage{tabularx,ragged2e,booktabs}
\usepackage{multirow}

\usepackage{color, colortbl}
\definecolor{Gray}{gray}{0.9}

\usepackage{enumitem}

\usepackage[heightadjust=all,valign=c]{floatrow}
\newfloatcommand{capbtabbox}{table}[][\FBwidth]

\usepackage[outline]{contour}
\usepackage{mathtools}
\usepackage{bbm}

\usepackage{sg-macros} 

\makeatletter
\renewcommand{\paragraph}{%
  \@startsection{paragraph}{4}%
  {\z@}{1.05ex \@plus 1ex \@minus .2ex}{-1em}%
  {\normalfont\normalsize\bfseries}%
}
\makeatother

\usepackage[breaklinks=true,bookmarks=false]{hyperref}

\iccvfinalcopy 

\def\iccvPaperID{****} 

\ificcvfinal\fi

\begin{document}

\title{An Alternative to WSSS? \\ An Empirical Study of the Segment Anything Model (SAM) on Weakly-Supervised Semantic Segmentation Problems}

\author{Weixuan Sun$^{1,2}$ \quad 
Zheyuan Liu$^{1}$ \quad
Yanhao Zhang$^{2}$\quad
Yiran Zhong$^{3}$ \quad
Nick Barnes$^{1}$\\ 
$^{1}$Australian National University \quad
$^{2}$OPPO Research Institute \quad
$^{3}$Shanghai AI Lab
}

\maketitle
\ificcvfinal\thispagestyle{empty}\fi

\begin{abstract}
The Segment Anything Model (SAM) has demonstrated exceptional performance and versatility, making it a promising tool for various related tasks. 
In this report, we explore the application of SAM in Weakly-Supervised Semantic Segmentation (WSSS). 
Particularly, we adapt SAM as the pseudo-label generation pipeline given only the image-level class labels.
While we observed impressive results in most cases, we also identify certain limitations. 
Our study includes performance evaluations on PASCAL VOC and MS-COCO, 
where we achieved remarkable improvements over the latest state-of-the-art methods on both datasets. 
We anticipate that this report encourages further explorations of adopting SAM in WSSS, as well as wider real-world applications.
The code is available at \url{https://github.com/weixuansun/wsss_sam}.


\end{abstract}

\section{Introduction}
Recent advancements in large foundation models have had a significant impact on the development of downstream deep-learning tasks. 
Large language models (LLMs) pre-trained on massive text corpus, \eg ChatGPT\footnote{https://openai.com/blog/chatgpt}
have demonstrated exceptional performance on a variety of Natural Language Processing~(NLP) tasks.
Concurrently, multi-modal pre-trained models including CLIP~\cite{radford2021learning} and BLIP~\cite{li2022blip} have been successfully applied to a range of vision and/or language tasks. 
The impact of these models is revolutionary in their respective domains.

The Segment Anything Model (SAM)~\cite{kirillov2023segment}
is a recent image segmentation model that has demonstrated outstanding performance in various applications.
SAM owes its impressive performances to two factors.
First, the model is trained on a large visual dataset named SA-1B containing over 1B masks of 11M images. 
This dataset is collected in a model-in-the-loop manner with multiple levels of human guidance.
The rich training data contributes to its ability to perform zero-shot segmentation on unseen images of varied distributions.
Second, SAM is designed and trained to accept versatile prompts as input.
To date, it supports points (via users clicking the mouse) or bounding boxes as prompts. This greatly improves the user experience when interacting with the model.

In this work, we focus on assessing the performance of SAM on image-level
Weakly-Supervised Semantic Segmentation (WSSS).
WSSS is initially proposed to reduce the label cost of fully supervised semantic segmentation~\cite{pathak2014fully}.
Instead of exhaustively labeling every pixel within an image for its class, WSSS resorts to weaker, yet cheaper alternatives. 
Examples include only labeling spare points~\cite{khoreva2017simple,bearman2016s}, bounding boxes~\cite{dai2015boxsup,sun20203d,song2019box,lee2021bbam}, or scribbles~\cite{vernaza2017learning,lin2016scribblesup,tang2018regularized} in an image, or using the image-level classification results as the label. The latter is the most common approach in recent work~\cite{wei2017object,jiang2019integral,zhang2020splitting,lee2021anti,sun2022inferring,xu2022multi,sun2021getam,Xie_2022_CVPR}, as such classification results can be readily obtained through any pre-trained vision backbones.
Pseudo segmentation labels are then extracted from these weak labels and are used to train the specific segmentation model that targets a certain domain or industrial application.
In this report, we aim to adapt SAM as the pseudo-label generation block in WSSS and examine its performance.

\begin{figure*}[htb!]
   \begin{center}
   {\includegraphics[width=1.0\linewidth]{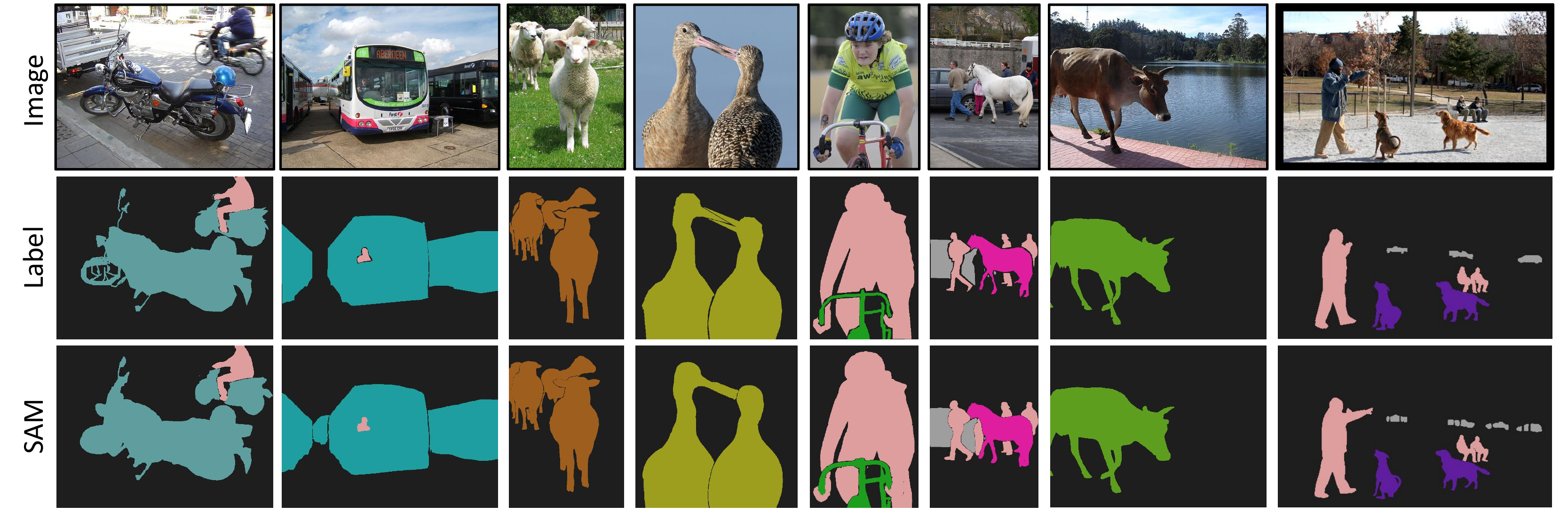}}
   \end{center}
\caption{SAM generated pseudo-labels compared to the ground-truth in PASCAL VOC. In most cases, SAM performs closely to the human annotations.}
   \label{fig: pascal pseudo}
\end{figure*}

\begin{figure*}[htb!]
   \begin{center}
   {\includegraphics[width=1.0\linewidth]{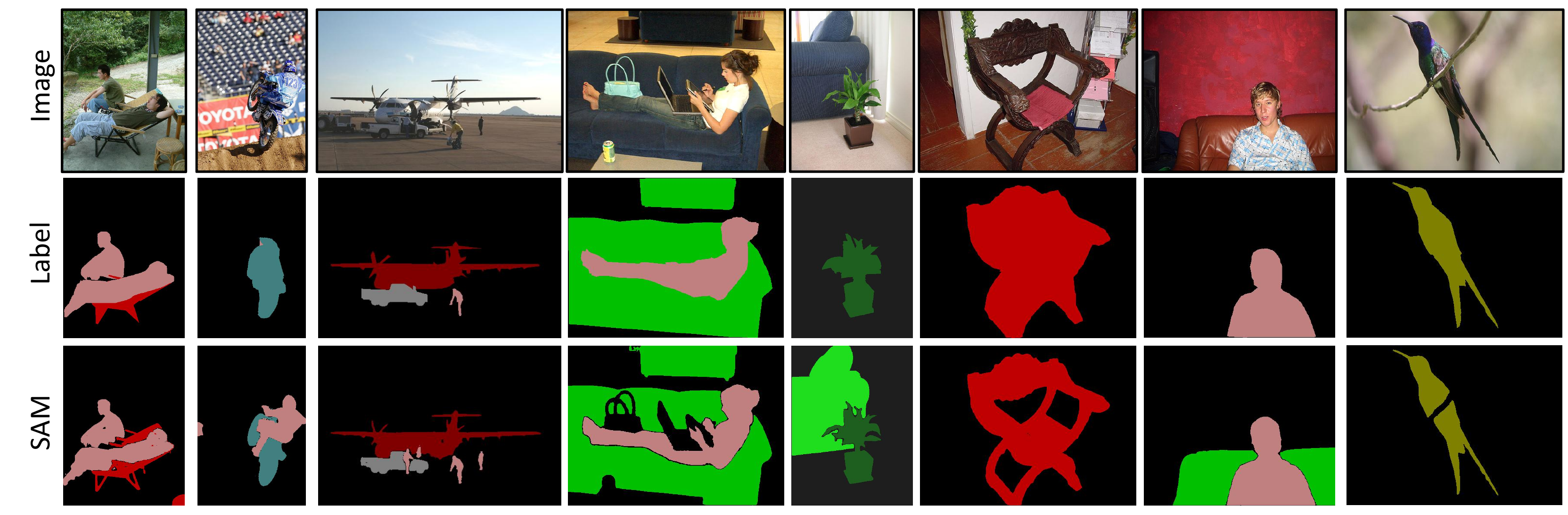}}
   \end{center}
\caption{We observe that in some cases SAM performs better than the human annotated ground-truth. Notably, SAM is able to capture crisp boundaries, more detailed structures and finer-grained semantic classes.}
   \label{fig: better pseudo}
\end{figure*}

\begin{figure*}[htb!]
   \begin{center}
   {\includegraphics[width=1.0\linewidth]{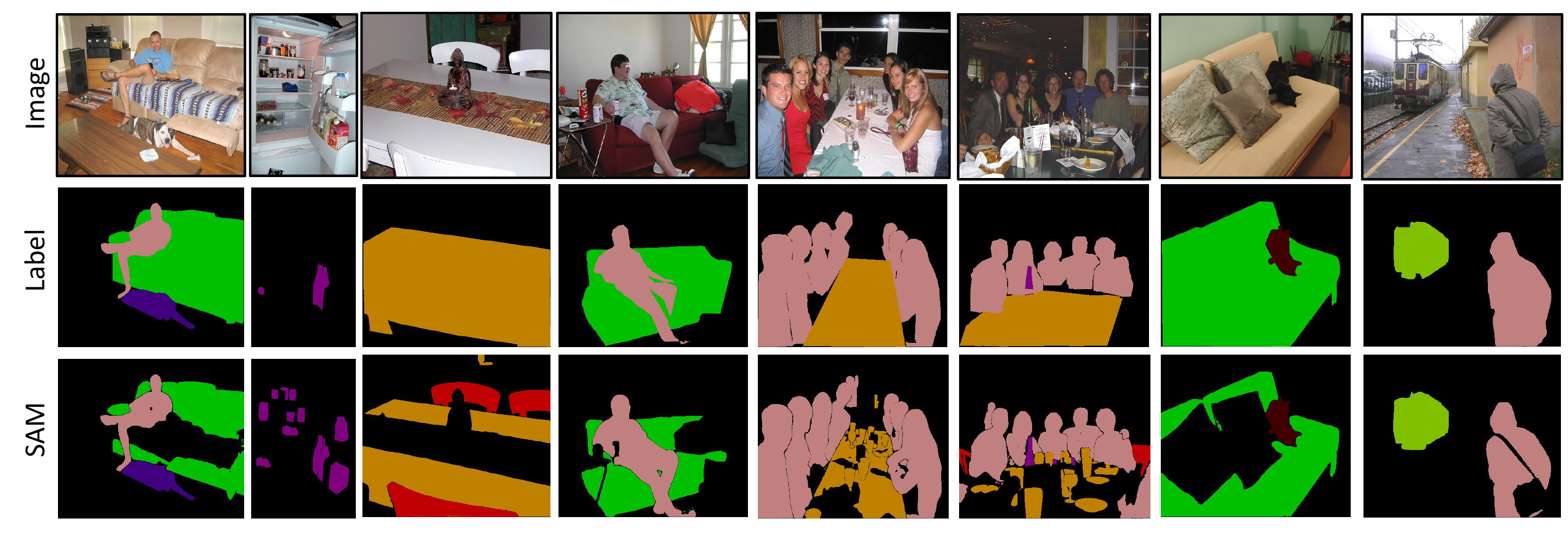}}
   \end{center}
\caption{Examples of semantic obscurity. SAM-generated segmentation masks may have different semantic granularity compared to human annotations. For example, in the second column, human labellers only consider two containers as ``bottles'' while SAM labels many more items as  ``bottles''.}
   \label{fig: issue}
\end{figure*}

We note that the training approach used for SAM includes pixel labeled images, which means that it cannot be regarded a weakly supervised approach. 
However, given the availability of a foundation approach for segmentation, we aim to compare the performances of traditional WSSS approaches with a foundation approach that is given the WSSS labels for the particular dataset. 
This is not a `fair' comparison, but we believe it is a practically valuable comparison.
Further, the Segment Anything paper states explicitly that at the start of the first training stage, ``SAM was trained using common public segmentation datasets". 
The paper does not give further information so the training may have had exposure to fully labeled images that are similar to some among the standard datasets for WSSS. 
However, given that the SAM paper evaluates a zero-shot instance segmentation on COCO, we infer that it is unlikely that COCO data was used for training.



We show quantitative and qualitative comparisons with state-of-the-art methods on image-level WSSS methods.
The main observation is that SAM performs favorably in most unprompted settings,
while it fails in certain cases due to the issue of semantic obscurity.
It suggests that SAM could potentially supersede the existing WSSS process as a practical approach. 
However, note that the question remains on how to effectively and efficiently apply it in real-world applications. 
Further, as stated, SAM does not qualify as a weakly-supervised approach.

\section{SAM for Weakly-Supervised Semantic Segmentation}
SAM is a powerful segmentation model comprised of a large image encoder, a prompt encoder, and a lightweight mask decoder. We refer readers to~\cite{kirillov2023segment} on architectural details.
The model currently supports point or bounding boxes as input prompts.
We empirically confirm that it performs well with such types of prompts, highlighting its ability in handling weak supervision in these formats.

However, to date, SAM does not support text as prompts --- the form of supervision used in image-level WSSS.
To this end, we enlist Grounded-Segment-Anything\footnote{https://github.com/IDEA-Research/Grounded-Segment-Anything} for our experiments on WSSS, which is an implementation that combines off-the-shelf image grounding methods with SAM.
Specifically, for every image, we concatenate the class labels in word(s) with full stops
as the text prompt and generate grounded bounding boxes via 
Grounded-DINO~\cite{ShilongLiu2023GroundingDM}.
Note that the text prompt is firstly tokenized into sub-words and grounding is then performed at the token level. 
Therefore, if a token (sub-word) is grounded in an image, we consider it as the grounding of its parent word. This approach enables us to map an arbitrary grounding back to one of the pre-defined class labels.
For example, class ``potted plant'' is tokenized into four tokens: ``pot'', ``\#ted'', ``\#pl'' and ``\#ant'', where the special character ``\#'' signifies that the corresponding token is an appending sub-word of the preceding token.
The bounding box is assigned to the class ``potted plant'' whenever any of the four tokens is grounded.
Once the grounded bounding boxes are generated, we feed them into SAM to produce instance masks.
Finally, we join the instance segmentation masks into semantic segmentation pseudo-labels.


\section{Experimental Results}
\subsection{Experimental setup}
\paragraph{Datasets} We apply SAM on two common WSSS datasets in our experiments, PASCAL VOC and MS-COCO.
PASCAL VOC has 10,582 training and 1,449 validation images.
MS-COCO is relatively larger-scaled, with 80k training and 40k validation images.
\paragraph{Implementation Details}
We use Grounded-DINO~\cite{ShilongLiu2023GroundingDM} to generate bounding boxes from text prompts with Swin-T as the grounding backbone.
For SAM, we use the ViT-H model to extract pseudo segmentation masks from bounding boxes. 
For the final semantic segmentation network, 
we use Deeplab-v2~\cite{chen2017deeplab} with ResNet-101~\cite{he2016deep} as backbone.
For MS-COCO, the model is pre-trained on ImageNet1k~\cite{deng2009imagenet}.
For PASCAL VOC, the model is pre-trained on COCO ground-truth following the setting of ~\cite{chen2017deeplab}.



\begin{table}[t!]
  \centering\scalebox{0.9}{
  \setlength{\tabcolsep}{2.5mm}
  \begin{tabular}{lc cc}
  \toprule
  Methods &  w/ saliency & Pseudo \\
  \midrule
  EDAM (CVPR2021)~\cite{wu2021embedded}  & \checkmark & 68.1 \\
  ReCAM (CVPR2022) \cite{chen2022class}   & \checkmark & 70.9 \\
  L2G (CVPR2022)~\cite{Jiang_2022_CVPR}   & \checkmark &71.9\\
  EPS (ECCV2020)~\cite{fan2020employing}     & \checkmark & 71.6 \\ 
  Du et al.(CVPR2022)~\cite{du2022weakly}   & \checkmark & 73.3  \\
  \hline
  PSA (CVPR2018)~\cite{ahn2018learning}  &  & 61.0  \\
  SEAM (CVPR2020)~\cite{wang2020self}  & & 63.6 \\
  CDA (ICCV2021)~\cite{su2021context}  & & 67.7 \\
  AdvCAM (CVPR2021)~\cite{lee2021anti}   & & 68.0 \\ 
  CPN (ICCV2021)~\cite{Zhang_2021_ICCV}  & & 67.8 \\
  Ru et al. (CVPR2022)~\cite{ru2022learning}  & & 68.7 \\
  Du et al.(CVPR2022)~\cite{du2022weakly}  &  & 69.2 \\
  MCTformer (CVPR2022)~\cite{xu2022multi}  &  &69.1 \\
CLIP-ES (CVPR2023)~\cite{lin2023clip} & & 75.0 \\ 
  \hline
  \textbf{SAM}    &   &     \textbf{88.3}            \\
  \bottomrule
  \end{tabular}
  }
  \caption{Performance of pseudo segmentation labels on the PASCAL VOC \textit{train} set. 
  SAM-based pseudo masks outperform previous methods by a significant margin. Best number is in bold.
  }
  \vspace{-3mm}
  \label{table pseudo}
  \end{table}

\begin{table}[t!]
  \caption{Performance comparison of WSSS methods on PASCAL VOC 2012 \textit{val} and \textit{test} sets. 
  w/ saliency: the method adopts extra saliency information. Best numbers are in bold.}
  \centering\scalebox{.85}{
  \setlength{\tabcolsep}{2.4mm}
  \begin{tabular}{lcccc}
  \toprule
  Methods & Venue & w/ saliency  & Val & Test  \\
  \midrule
  NSRM~\cite{yao2021non}    & CVPR2021 & \checkmark   & 70.4    &  70.2\\
  InferCam~\cite{sun2022inferring} & WACV2022 & \checkmark & 70.8 & 71.8 \\
  EDAM~\cite{wu2021embedded} &CVPR2021 &\checkmark & 70.9 & 70.6 \\
  EPS~\cite{lee2021railroad} &CVPR2021& \checkmark & 71.0 & 71.8 \\
  DRS~\cite{kim2021discriminative} &AAAI2021& \checkmark & 71.2 & 71.4 \\
  L2G~\cite{Jiang_2022_CVPR} &CVPR2022& \checkmark & 72.1 & 71.7 \\
  Du et al. ~\cite{du2022weakly}&CVPR2022 & \checkmark & 72.6 & 73.6 \\
  \hline
  PSA~\cite{ahn2018learning} & CVPR2018&  &61.7  & 63.7  \\
  SEAM~\cite{wang2020self} &CVPR2020 &  & 64.5 & 65.7 \\
  CDA~\cite{su2021context}&ICCV2021 & & 66.1 & 66.8 \\
  ECS-Net~\cite{Sun_2021_ICCV}&ICCV2021 &&66.6 & 67.6 \\
  Du et al. ~\cite{du2022weakly}&CVPR2022 & &67.7 & 67.4\\
  CPN~\cite{Zhang_2021_ICCV}&ICCV2021  &    & 67.8   &  68.5\\
  AdvCAM~\cite{lee2021anti}&CVPR2021 & & 68.1 & 68.0 \\  
  Kweon et al.~\cite{Kweon_2021_ICCV}&ICCV2021 && 68.4 &68.2 \\
  ReCAM~\cite{chen2022class}&CVPR2022  &   &  68.5 & 68.4 \\
  SIPE~\cite{Chen_2022_CVPR}& CVPR2022 & & 68.8 & 69.7 \\
  URN~\cite{li2021uncertainty}&AAAI2022 & & 69.5 & 69.7 \\
  ESOL~\cite{li2022expansion} & NeurIPS2022 & & 69.9 & 69.3 \\
  PMM~\cite{Li_2021_ICCV}&ICCV2021  &    & 70.0    &  70.5\\
  VWL-L~\cite{ru2022weakly}&IJCV2022  &   & 70.6 & 70.7 \\
  Lee et al.~\cite{lee2022weakly} & CVPR2022 & & 70.7 & 70.1 \\
  MCTformer ~\cite{xu2022multi} & CVPR2022 & & 71.9 & 71.6 \\
  OCR~\cite{cheng2023outofcandidate} & CVPR2023 && 72.7 & 72.0 \\
  CLIP-ES~\cite{lin2023clip} & CVPR2023 && 73.8 & 73.9 \\
    \hline
  \textbf{SAM}   &   &  &\textbf{77.2} &    \textbf{77.1}      \\
  \hline
  full-supervision~\cite{chen2017deeplab} & & &  77.7 & 79.7 \\
  \bottomrule
  \end{tabular}}
  \label{table pascal seg}
  \end{table}

\begin{table}[t!]
  \caption{Performances of pseudo segmentation labels on MS COCO \textit{train} set. 
  SAM pseudo masks outperform previous methods by a significant margin. Best number is in bold.
  }
  \centering\scalebox{0.9}{
  \setlength{\tabcolsep}{2.5mm}
  \begin{tabular}{lc cc}
  \toprule
  Methods &  w/ saliency & Pseudo \\
  \midrule
  AdvCAM (CVPR2021)~\cite{lee2021anti}   & & 35.8 \\ 
  IRN (CVPR2018)~\cite{ahn2019weakly}  &  & 42.4  \\
  ReCAM (CVPR2022) \cite{chen2022class}   & \checkmark & 46.3 \\
  \hline
  \textbf{SAM}    &   &     \textbf{66.8}            \\
  \bottomrule
  \end{tabular}
  }
  \vspace{-3mm}
  \label{table coco pseudo}
  \end{table}

  \begin{table}[t!]
    \caption{Segmentation performance comparison of WSSS methods on MS COCO. w/ saliency: the method adopts extra saliency information.
    Best number is in bold.}
    \centering\scalebox{.85}{
    \setlength{\tabcolsep}{3.2mm}
    \begin{tabular}{lccc}
    \toprule
    Methods & Venue & w/ saliency  & Val  \\
    \midrule
    AuxSegNet \cite{Xu_2021_ICCV}& ICCV2021 & \checkmark & 33.9 \\
    EPS~\cite{fan2020employing} &CVPR2022& \checkmark &  35.7 \\
    L2G~\cite{Jiang_2022_CVPR}&CVPR2022 & \checkmark & 44.2 \\
    \hline
    Wang et al.~\cite{wang2018weakly}&IJCV2020 & & 27.7 \\
    Ru et al.~\cite{ru2022learning}&CVPR2022 &  & 38.9  \\
    SEAM~\cite{wang2020self}&CVPR2020 && 31.9 \\
    CONTA~\cite{zhang2020causal}&NeurIPS2020 && 32.8  \\
    CDA~\cite{su2021context}&ICCV2021 & & 33.2  \\
    Ru et al.~\cite{ru2022weakly} & IJCV2022 & & 36.2 \\
    URN~\cite{li2021uncertainty}&AAAI2022  && 41.5 \\
    MCTformer~\cite{xu2022multi} &CVPR2022 &  & 42.0 \\
    OCR~\cite{cheng2023outofcandidate}&CVPR2023 & & 42.5 \\
    ESOL~\cite{li2022expansion} & NeurIPS2022 & &42.6 \\
    SIPE~\cite{Chen_2022_CVPR}&CVPR2022 & & 43.6 \\
    RIB~\cite{lee2021reducing}&NeurIPS2020 & & 43.8 \\
    CLIP-ES~\cite{lin2023clip}&CVPR2023 && 45.4 \\
    \hline
    \textbf{SAM}  & &   & \textbf{55.6}          \\
    \bottomrule
    \end{tabular}
    }
    
    \label{table coco}
    \end{table}

\section{Quantitative Analysis}
\subsection{PASCAL VOC}
We present pseudo-label and final segmentation performances on PASCAL VOC in Table~\ref{table pseudo} and ~\ref{table pascal seg} respectively. 
As shown in Table~\ref{table pseudo}, our pseudo segmentation masks achieve 88.3 mIoU, which outperforms previous state-of-the-art by 13.3 mIoU.
It validates that SAM generates segmentation masks of extremely high quality for the targeted classes.
In Table~\ref{table pascal seg}, SAM-based results achieve 77.2 mIoU on \textit{val} set and 77.1 mIoU on \textit{test} set, both surpassing previous state-of-the-art methods with clear margins. 
We note that the segmentation performances are approaching the fully-supervised results on PASCAL VOC.

\subsection{MS-COCO}
We show pseudo-label and segmentation performance comparison on MS-COCO in Table~\ref{table coco pseudo} and Table~\ref{table coco}.
Likewise in the PASCAL VOC,
The SAM-based method achieves an appreciable segmentation mIoU of 55.6, which surpasses existing methods immensely. 
Compared to PASCAL VOC, MS-COCO is a lager dataset with more semantic classes and complex images that include multiple objects, this encouraging result further demonstrates that SAM could handle annotations of complex scenes.


\section{Discussion}
\subsection{Why Don't We Directly Use SAM?}
One might question the need for a Weakly Supervised Semantic Segmentation (WSSS) pipeline given the impressive performance of SAM. However, we argue that a WSSS pipeline is still beneficial for certain application scenarios.
For instance, in many narrow-domain use cases, only specific semantic classes are of interest, which renders the open-vocabulary setup unnecessary and even prone to introducing errors. 
In contrast, a WSSS pipeline can be trained to respond to only a selected set of classes, which is more favored.
Moreover, industrial or mobile environments that are often resource-limited and time-sensitive cannot accommodate the use of SAM, with its considerable VRAM usage and low inference speed. 
In such cases, training a specific semantic segmentation network in the WSSS fashion for the downstream tasks remains meaningful. 



\subsection{Pseudo Label Quality}
SAM can generate high-quality segmentation masks as shown in Fig.~\ref{fig: pascal pseudo}.
In addition,
we observe that the pseudo-labels generated via SAM are more accurate than the manual annotations in some cases.
As shown in Fig.~\ref{fig: better pseudo}, human annotations often contain imprecise polygons and ignore fine-grained details.
In contrast, SAM is able to capture crisp boundaries, more detailed structures, and finer-grained semantic classes.
We conjecture the possibility that SAM can be used to guide, or even correct human annotations when building future datasets.

\subsection{The Issue of Semantic Obscurity}
When generating pseudo-labels, we encounter the issue of ``semantic obscurity'', which arises from the subjectivity of human annotations. 
A typical example is shown in Fig.~\ref{fig: issue}, where ground-truth annotations in PASCAL VOC for ``dining table'' always include all objects that are placed on the table --- ``plate'', ``bowl'', and ``food'' etc.
However, the output of SAM stays true to the concept of ``table'' and contains nothing else.
We suspect the issue is caused by the discrepancy between PASCAL VOC and the SAM SA-1B dataset, particularly the granularity of the annotations.

Granted, one could argue that the PASCAL VOC dataset is flawed in this regard.
Nevertheless, to assess the true performance of SAM on said dataset, we attempt to relieve this issue by including some assist words, such as ``bowl'', ``plate'', ``food'', ``fruit'', ``glass'', and ``dishes'' for the affected ``dining table'' class; 
and likewise, ``halter'' and ``saddle'' for the ``horse'' class. 
We derive these words through manual examination. Note that this mitigation strategy is by no means comprehensive, nor generalizable to datasets of larger scales (\eg MS-COCO), as it might interfere with other classes.
A systematic approach to addressing the semantic obscurity issue would be to construct hierarchical-structured semantic classes with better prompts. We leave this to future work.

\subsection{Conclusion and Future Work}
This report presents a preliminary investigation into the application of SAM as a foundation model in Weakly Supervised Semantic Segmentation (WSSS).
Results from experiments conducted on two datasets demonstrate that SAM can yield competitive performance without the need for model fine-tuning, 
while for existing WSSS methods, the classification re-training and pseudo-label generation are burdensome necessities. 
Notably, we are not aiming to achieve SOTA results in a fair comparison, as SAM is trained with massive amounts of data and a training process that includes fully labeled images.
This research concentrates on the direct application of the SAM as a foundation model, which enables a streamlined and simplified approach to WSSS.

Moving forward, 
a potential direction would be to adapt SAM into an automatic labeling tool for various real-world applications. 
Specifically, we shall explore the use of hierarchical-structured semantic classes and better prompts, which address the issue of semantic obscurity. 
We might also wish to investigate using SAM for segmenting ``stuff'' (i.e., non-object, background) classes such as ``sky'', ``sea'' and ``road'',  in order to improve the overall scene understanding ability.

{\small
\bibliographystyle{ieee_fullname}
\bibliography{egbib}
}

\end{document}